\documentclass{article}
\usepackage{ijcai22}

\usepackage{times}
\usepackage{soul}
\usepackage{url}
\usepackage[hidelinks]{hyperref}
\usepackage[utf8]{inputenc}
\usepackage[small]{caption}
\usepackage{graphicx}
\usepackage{amsmath}
\usepackage{amsthm}
\usepackage{booktabs}
\usepackage{algorithm}
\usepackage{algorithmic}
\usepackage{subfigure}
\usepackage{multirow}
\usepackage{booktabs}
\usepackage{array}
\usepackage{makecell}

\usepackage{cleveref}
\title{RMBR: A Regularized Minimum Bayes Risk Reranking Framework \\ for Machine Translation}

\author{
Yidan Zhang$^1$
\and
Yu Wan$^2$\and
Dayiheng Liu$^1$\and
Baosong Yang$^2$\and
Zhenan He$^1$
\affiliations
$^1$College of Computer Science, Sichuan University \\
$^2$NLP$^2$CT Lab, University of Macau\\
\emails
losinuris@gmail
}

\begin{document}

\maketitle

\begin{abstract}
Beam search is the most widely used decoding method for neural machine translation (NMT). In practice, the top-1 candidate with the highest log-probability among the $n$ candidates is selected as the ‘preferred’ one. However, this top-1 candidate may not be the best overall translation among the $n$-best list. Recently, Minimum Bayes Risk (MBR) decoding has been proposed to improve the quality for NMT, which seeks for a consensus translation that is closest on average to other candidates from the $n$-best list. We argue that MBR still suffers from the following problems: The utility function only considers the lexical-level similarity between candidates; The expected utility considers the entire $n$-best list which is time-consuming and inadequate candidates in the tail list may hurt the performance; Only the relationship between candidates is considered. To solve these issues, we design a regularized MBR reranking framework (RMBR), which considers semantic-based similarity and computes the expected utility for each candidate by truncating the list. We expect the proposed framework to further consider the translation quality and model uncertainty of each candidate. Thus the proposed quality regularizer and uncertainty regularizer are incorporated into the framework. Extensive experiments on multiple translation tasks demonstrate the effectiveness of our method.
\end{abstract}

\section{Introduction}\label{sec:intro}
Given a source sentence, neural machine translation (NMT) \cite{55} models are trained to predict conditional probability distributions for candidate translations. In practice, it is desirable to output a single sentence, not a distribution. Therefore, a decision rule is required to rank the candidates and select the ‘preferred’ one. The most widely used decision rule is maximum-a-posteriori (MAP) decoding, which seeks the most probable translation under the conditional distribution. Due to the huge search space, beam search is proposed as an approximation. Given a pre-defined beam size $n$, beam search always keeps the top-$n$ candidates based on the log-probability score. Then, the top-1 candidate, \textit{i.e.}, the one with the highest log-probability among the $n$-best list, is selected as the ‘preferred’ one. Unfortunately, this top-1 candidate might not be the best translation on the $n$-best list. 

\begin{figure*}[htbp]
	\centering 
	\subfigure[]{
		\label{Fig.sub.1}
		\includegraphics[width=0.34\textwidth]{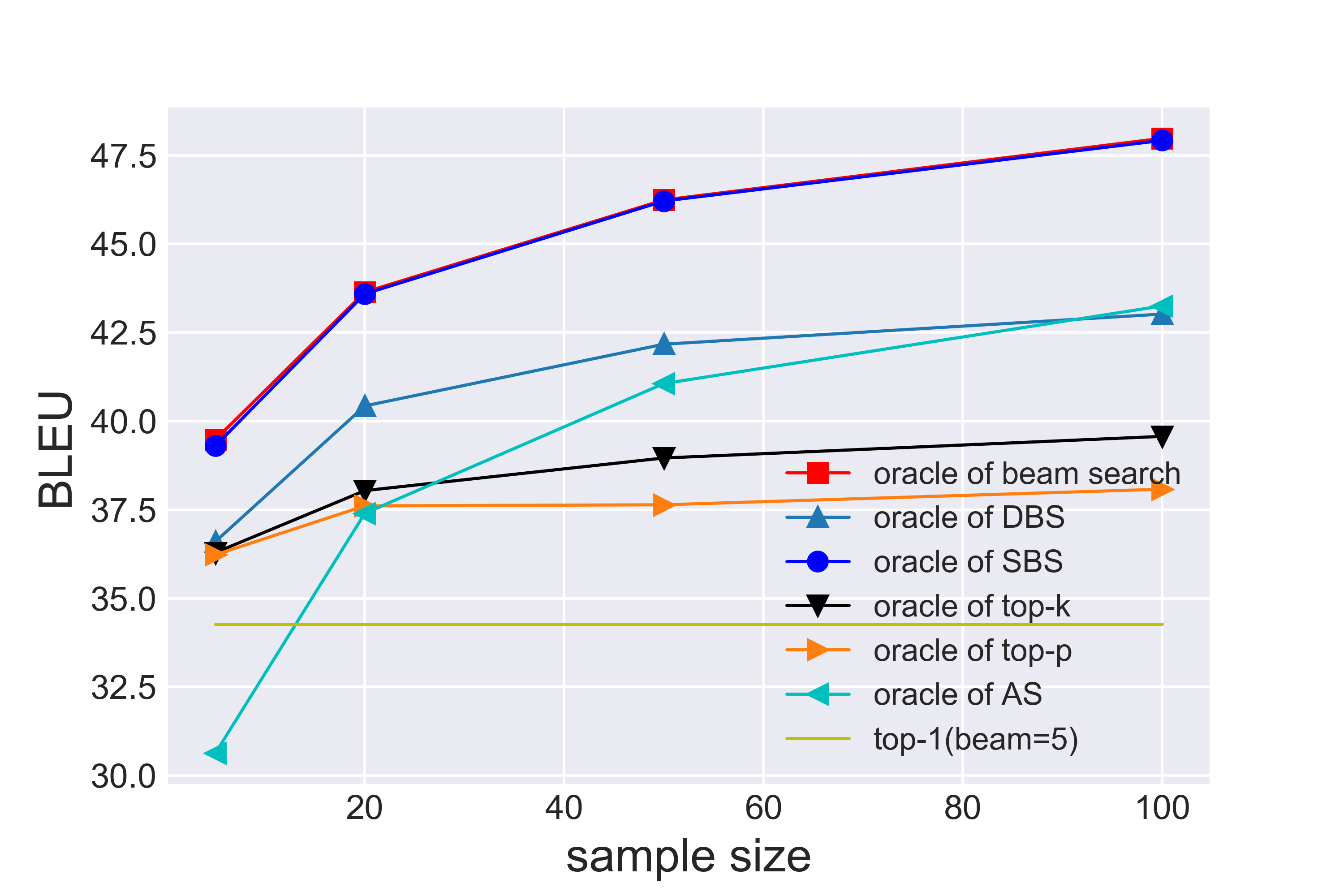}}
	\hspace{-0.80cm}
	\subfigure[]{
		\label{Fig.sub.3}
		\includegraphics[width=0.34\textwidth]{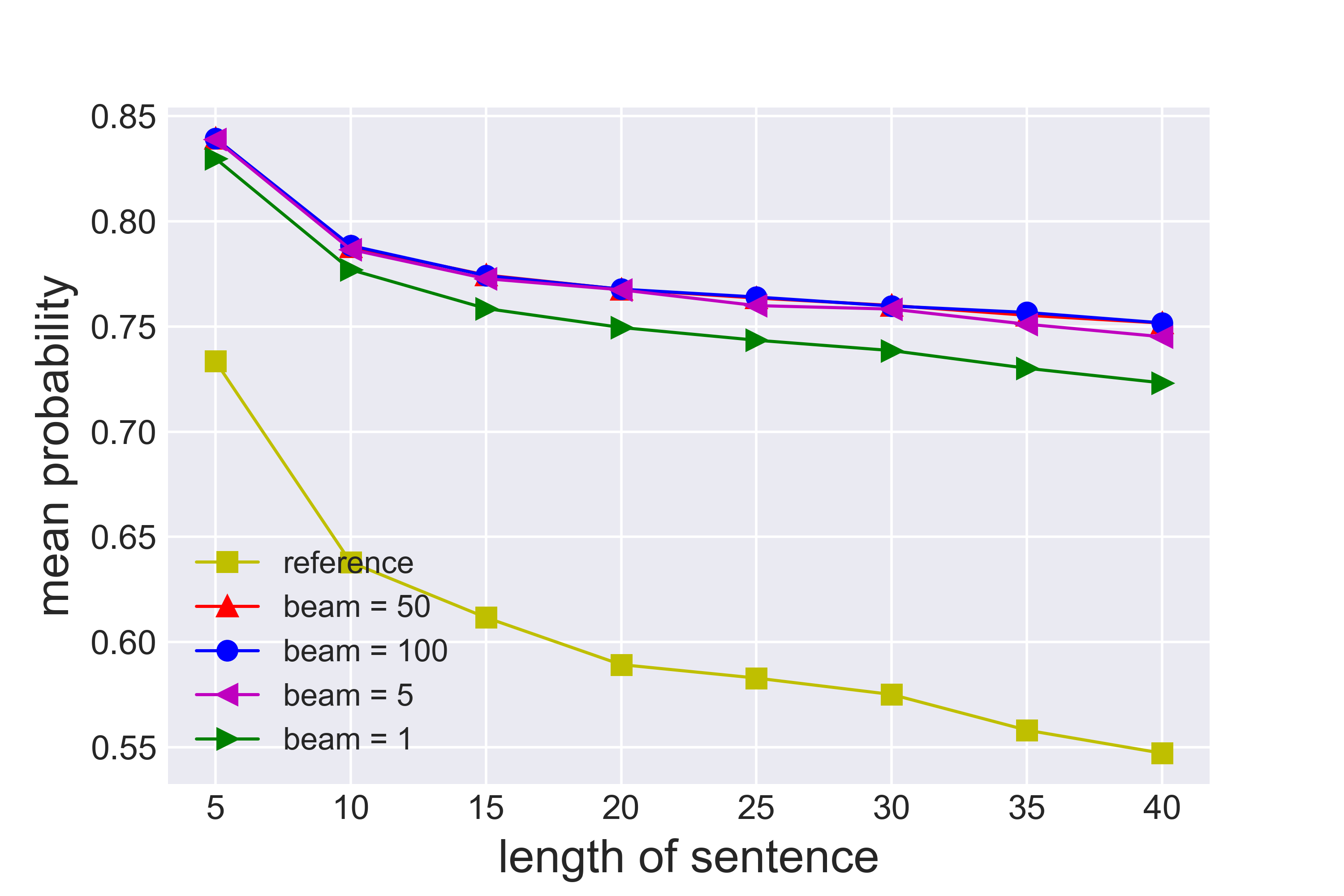}}
	\hspace{-0.80cm}
	\subfigure[]{
		\label{Fig.sub.2}
		\includegraphics[width=0.34\textwidth]{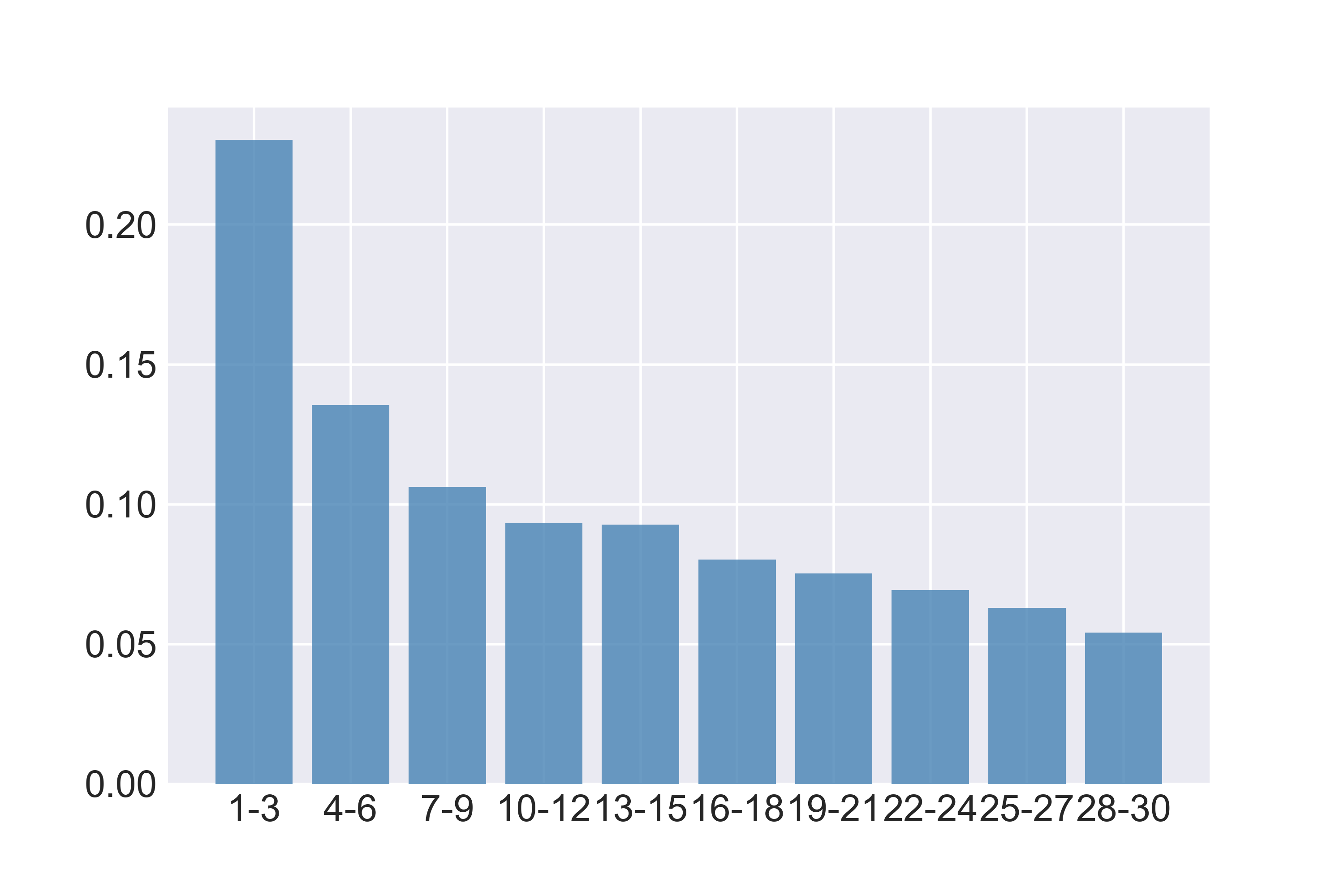}}
	\caption{An example of exploring candidate spaces on the IWSLT'14 De$\to$En test set. (a) Oracle ranking of samples generated by multiple decoding strategies. (b) The token probabilities of sentences in different length intervals. The x-axis is the length interval, and the y-axis is the average token probability of the sentences within the same length range. (c) The distribution of oracle translations’ rank index in the $n$-best list. The x-axis represents the index interval, and the y-axis represents the proportion of oracle translations indexed in an interval.}
	\label{f1}
\end{figure*}

We conduct oracle experiments to explore the performance gap between the oracle result\footnote{The oracle result is defined as $\text{argmax}_{Y\sim p_{\text{NMT}(Y|X)}}$ BLEU$(Y, Y')$, where $(X, Y')$ is the pair of source and reference sentence.} in the $n$-best candidates and top-1 candidate. Besides using beam search, we further use three stochastic decodings (ancestral search (AS) \cite{8}, top-$k$ \cite{9}, top-$p$ \cite{10}), and two deterministic decodings (diverse beam search (DBS) \cite{11}, sibling beam search (SBS) \cite{12}) to obtain $n$ candidates, respectively. The results are reported in Fig. \ref{f1}a. Overall, all of the oracle results achieve \textit{significantly} higher BLEU scores than the top-1 candidate of beam search with beam size 5. %baseline (the results by running beam decoding with a beam size of five). 
%For example, under the beam size 100, an oracle result of beam search achieves the high BLEU score of 47.98, while the top-1 candidate achieves only 34.28. 
Furthermore, we observe that under the oracle experiment, using beam search to obtain $n$-best candidates still outperforms other decoding methods. These results suggest that beam search actually performs well, yet log-probability scores fail to select the best translation from the $n$-best list. Similar to our study, ~\citeauthor{20}~\shortcite{20} has observed that NMT model is capable of outputting high-quality candidate translations, but fails at picking them as the best one. ~\citeauthor{5}~\shortcite{5} also points out that, NMT models are good at spreading probability mass over a large number of acceptable outputs, but they are not efficient at selecting the best one. 

To further explore why the top-1 candidate is not the best translation, we compare the token probability between top-1 candidates and references. Specifically, the average probability of all the tokens in each sentence is firstly computed, which is defined as the token probability. To eliminate the effect of sentence length, the mean token probability of all candidates in the same length range is observed. As shown in Fig. \ref{f1}b, we find that the token probability of top-1 candidates is much higher than that of references, especially when the result length is longer, suggesting that NMT models may over-confident about the top-1 candidates. During beam search decoding, assigning an excessively high probability to a suboptimal sequence in one step can lead to a chain reaction that eventually produces an unnatural candidate with high probability. Besides, we argue that the essence of the beam search curse \cite{45} (large beam sizes hurt translation quality) is lying in the token probability gap between top-1 candidates and reference translations, as larger beam sizes lead to larger gaps from Fig. \ref{f1}b.

In view of the above analysis, we expect to find a consensus candidate from the $n$-best list to avoid the ``over-confident'' candidates. %Recently, a new decision rule, Minimum Bayes Risk (MBR) decoding, is proposed in NMT to find the translation that is closest to other candidate translations to minimizes the expected risk for a given utility function.
Recently, a new decision rule, Minimum Bayesian Risk (MBR) decoding, has been proposed in NMT. The main idea of this method is to find the translation that is closest to other candidate translations to minimize the expected risk for a given utility function.
In \citeauthor{14}~\shortcite{14} and \citeauthor{20}~\shortcite{20}, MBR decoding are combined with beam search to improve the translation quality. Nevertheless, we argue that there are still some defects in MBR decoding: (a) The utility function only considers the lexical-based similarity between candidates, such as BLEU, METEOR, CHRF etc.; (b) The expected utility for each candidates considers the entire $n$-best list, which requires a large computational cost, especially when $n$ is large. Besides, inadequate candidates in the tail list may hurt the performance; (c) MBR only considers the similarity between candidates but completely ignore the model uncertainty and the translation quality of each candidate.

To solve above issues, we propose a \textbf{R}egularized \textbf{M}mini-mum \textbf{B}ayesian \textbf{R}isk reranking framework (\textbf{RMBR}). For the first problem, we explore the use of semantic-based evaluation metrics (\textit{e.g.}, COMET \cite{27} and BLEURT \cite{28}) as the utility function for MBR. Aiming at the second issue, we conduct experiment to analyze the probability ranking of the oracle translations in the $n$-best list ($n$=30). As shown in Fig. \ref{f1}c, the oracle translations are less likely to appear in the tail list. Therefore, we use only the top-$l$ $(l \leq n)$ candidates of the $n$-best list to calculate the MBR score (expected utility) for each candidate. In this way, the computational cost is reduced and the inadequate candidates in the tail list that is close to each other, are avoided. For the third problem, we incorporate two types of regularizers into the framework: quality regularizer and uncertainty regularizer. Quality regularizer allows RMBR framework to further consider the translation quality of a single candidate in addition to considering the similarity between candidate results. To be concrete, we consider four regularization scores as the quality regularizer: language model score, back-translation score~\cite{52}, quality estimation score, and translation score (log-probability score). While the uncertainty regularizer aims to further consider the model uncertainty for each output. In this paper, we explore two kinds of uncertainty regularizers: Monte Carlo (MC) Dropout \cite{53,35} and the entropy of model output distributions.

%Quality estimation scores \cite{50}, language model scores \cite{51}, and back-translation scores \cite{52} have been experimentally demonstrated the effectiveness on NMT evaluation, thus can be used as the quality regularizer. In addition, we design another uncertainty regularizer based on uncertainty estimation to rerank the $n$-best list avoiding over-confident candidates with unnatural high probability. First, Monte Carlo (MC) Dropout \cite{53,35} is utilized to explore candidates with stable posterior probabilities under perturbation. Second, entropy measures are utilized to explore candidates with a uniform token probability distribution. 
%In summary, our contributions are as follows: (1)
We conduct extensive experiments to compare different settings of RMBR, as well as the previous MBR method~\cite{14,20} using BLEU as utility and several commonly used translation reranking methods. Experimental results show that after using COMET as utility function, our MBR outperforms previous MBR decoding methods~\cite{14,20}. When the proposed quality regularizer or uncertainty regularizer is further introduced, the performance of RMBR can be further improved. Our method achieves consistent performance gains on the tasks of German-English from IWSLT’14, and German-English, English-German, and English-French tasks from WMT’14, which demonstrates the effectiveness of our method.

%Our main contributions can be summarised as follows:
%\begin{itemize}
%	\item We observe that using partial candidates of the $n$-best list to compute expected utility aids to improvement on BLEU score.
%	\item We design uncertainty and quality regularizes. beam size note or not?
%	\item We propose a regularized MBR reranking framework and empirically demonstrate that it can improve upon beam search.
%\end{itemize}

\section{Preliminary}
\subsection{The Decoding Problem}
Let $X=\{x_1,x_2,...,x_{|X|}\}$ denote a source sequence, $Y=\{y_1,y_2,...,y_{|Y|}\}$ denote a target sequence. A NMT model defines a distribution over outputs and sequentially predicts tokens using a softmax function as follows:
\begin{equation}
p(Y|X)=\prod_{t=1}^{|Y|}p\textsubscript{NMT}(y_t|X,y_1,y_2,...,y_{t-1}).
\end{equation}
The decoding problem can be written as finding a sequence $Y^\ast$ that maximizes the probability given input $X$:
\begin{equation}
Y^\ast= \mathop{\arg\max}\limits_{X} p\textsubscript{NMT}(Y^\ast|X).
\end{equation}

\subsection{Beam Search}
When decoding with the above distribution over sequences, it is not feasible to pick out the most probable sequence among all possible sequences. A common approximate decoding method is beam search, which maintains the top-$n$ highly scoring candidates at each time step. $n$ is known as beam size, and the log-probability of a sequence at time $t$ is computed as:
\begin{equation}
S(Y_{t}|X)= S(Y_{t-1}|X)+\text{log } p\textsubscript{NMT}(y_t|X,Y_{t-1}),
\end{equation}
where $S(Y_{t-1}|X)=\text{log } p\textsubscript{NMT}(y_1,y_2,...,y_{t-1}|X)$. The decoding process is repeated until the stop condition is met. After that, we can obtain a list of $n$ most promising candidates. Finally, the most likely sequence is selected as the ‘preferred’ translation by ranking the $n$ candidates based on log-probability scores $S(Y|X)$. 

\section{Regularized MBR Reranking Framework}
As discussed in Sec~\S\ref{sec:intro}, picking the candidate with the highest log-probability score is unable to effectively obtain the best result. %In particular, simply selecting , which is formally defined as the top-1 candidate, may widen the gap between references. 
In this paper, we propose a regularized MBR reranking framework (RMBR) that adopts the semantic similarity evaluation metric as the utility function. Besides considering the similarity between the output candidates, we expect the proposed framework to further consider the translation quality of each candidate and the uncertainty of the model. Thus we incorporate two types of regularizers into the framework: Quality Regularizer (Sec~\S\ref{sec:QR}) and Uncertainty Regularizer (Sec~\S\ref{sec:UR}). The candidate with the highest reranked score is formally defined as the 1-best candidate.

%The goal of the proposed method is to score and select the candidate in an overall way. 
Given a list of $n$ most likely candidates generate by beam search with beam size $n$, which can be written as $\{H_1,H_2,...,H_n\}$. %and sorted in descending order of log-probability scores.
The regularized score for $H_i$ is computed as:
\begin{equation}
S\textsubscript{RMBR}(H_i)=S\textsubscript{MBR}(H_i)+\sum \lambda_j \mathcal{R}_j(H_i),
\end{equation}
where $S\textsubscript{MBR}$ is the MBR score, which is introduced in the next section. Note that we introduce two types of regularizers, $\mathcal{R}_j$ is used to denote the $j$-th regularizer score. $\lambda_j$ is a tradeoff parameter\footnote{$\lambda_j$ is selected from the set \{0.001, 0.01, 0.1, 1, 10\} with the best performance on the validation set.} to achieve a satisfying balance among multiple decoding objectives. Finally, the 1-best candidate is selected as the ‘preferred’ translation. 

%The chosen of $R(\cdot)$ can be the score of sentence-level automatic evaluation methods, and 

\subsection{MBR Score}
\label{MBR Score}
Given a utility function $\mathcal{U}$ (\textit{e.g.}, BLEU) and a list of $n$-best candidates. The MBR score (expected utility) for each candidate is computed by comparing it to all other candidates in the $n$-best list. Since only a few oracle translations appear at the tail list as we observed in preliminary experiment, we compute the MBR score for $H_i$ by comparing it to top-$l$ candidates:
\begin{equation}
S\textsubscript{MBR}(H_i)= \frac{1}{l} \sum_{j=1}^{l} \mathcal{U}(H_i,H_j),
\end{equation}
where $l\in \{1,2,...,n\}$ is tuned on the validation set and fixed for inference for all testing instances. The candidate with the highest MBR score $S\textsubscript{MBR}$ is the consensus translation in the $n$ candidates. Besides using lexical-based method (BLEU) as utility function $\mathcal{U}$ which is called MBR\textsubscript{BLEU}, we further explore two semantic-based evaluation methods BLEURT and COMET as utility functions $\mathcal{U}$ in our framework, which are called MBR\textsubscript{BLEURT} and MBR\textsubscript{COMET}, respectively.

\subsection{Quality Regularizer}\label{sec:QR}
MBR score only considers the similarity between the output candidates and ignores the translation quality of each candidate. To bridge this gap, we introduce a quality regularizer into MBR framework. In this work, we explore four kinds of scores as the quality regularizer: a) Language Model (LM) score; b) Back-Translation (BT) score; c) Quality Estimation (QE) score; and d) log-probability scores. The computation for candidate $H_i$ is as follows:
\begin{align}
\text{LM}(H_i)&=\text{log} p\textsubscript{LM}(H_i), \text{BT}(H_i)=\text{log} p\textsubscript{NMT}(X|H_i), \\
\text{QE}(H_i)&=f\textsubscript{QE}(X,H_i),
\end{align}
where log$p\textsubscript{LM}(H_i)$ is calculated by a pre-trained language model, $p\textsubscript{NMT}(X|H_i)$ is via a backward NMT model, and $f\textsubscript{QE}(X,H_i)$ is by a off-the-shelf quality estimation model (\textit{e.g.}, TransQuest~\cite{50}).

\subsection{Uncertainty Regularizer}\label{sec:UR}
In this section, we introduce the uncertainty regularizer, which quantifies whether the current model is confident or hesitant on the candidate translation. For efficiency, we utilize widely used Monte Carlo (MC) dropout and entropy measures to compute model uncertainty.

\noindent\textbf{MC Dropout.} At test time, for a candidate $H_i$ paired with input $X$, we perform $m$ forward passes through the NMT model parameterized by $\hat{\theta}$, where the $t$-th pass randomly deactivates part of neurons. Then, $m$ sets of sentence-level perturbed log-probability score are collected, which is written as:
\begin{equation}
\text{MC}_{\hat{\theta}_t}(H_i)=-\text{log }p\textsubscript{NMT}(H_i|X,\hat{\theta}_t).
\end{equation}

\noindent\textbf{Entropy Measures.} We also consider using the entropy of model predicting probability distribution of each candidate as a measure of model uncertainty. Intuitively, given an output sample, if the model probability distribution entropy of each token is very small, it means that the model has a high degree of confidence in this output result. Let $\mathcal{V}=\{v_1,v_2,...,v_{|V|}\}$ denote the target vocabulary of NMT, we compute the token entropy for each token in the candidate $H_i=\{h_{i1},h_{i2},...,h_{i|H_i|}\}$. Then $|H_i|$ sets of token entropy are collected, which is written as:
\begin{equation}
S\textsubscript{entropy}(h_{it})=-\sum_{j=1}^{|\mathcal{V}|}\text{log}p\textsubscript{NMT}(v_j|X,h_{i1}, ..., h_{it-1}).
\end{equation}

Finally, the expectation of $m$ sets of $\text{MC}_{\hat{\theta}_m}(H_i)$ and $|H_i|$ sets of $S\textsubscript{entropy}(h_{it})$ are used as the uncertainty regularizer score.

\begin{table*}
\centering
\scalebox{0.8}{
\begin{tabular}{lccc|ccc}
\\ \toprule
& \multicolumn{3}{c}{IWSLT’14 De$\to$En}        & \multicolumn{3}{c}{WMT’14 De$\to$En} \\ \hline
Method           & COMET & BLEURT & BLEU     & COMET    & BLEURT & BLEU    \\ \hline
Top-1 (beam=5)   & 34.79 & 16.16  & 34.28 & 42.35     & 21.90      & 32.70    \\ 
Top-1 (beam=30)  & 34.22 & 15.99  & 34.17 & 41.80     & 21.60      & 32.54    \\ \hline
LP+BT \cite{52}  & 40.63 & 18.57  & 35.11 & 45.94     & 23.42      & 33.06    \\
LP+QE \cite{50}  & 38.84 & 19.53  & 35.37 & 45.56     & 24.30      & 33.41    \\
LP+LM \cite{51}  & 36.33 & 16.58  & 35.14 & 44.48     & 22.48      & 33.49    \\ 
Range Voting \cite{29}           & 34.89 & 16.59  & 34.53 & 42.29    & 21.53     & 32.78    \\
MBR\textsubscript{BLEU}(full)~\cite{20}          & 33.76 & 15.91  & 34.38 & 41.66     & 20.96      & 32.68    \\ \hline
MBR\textsubscript{BLEU}           & 34.39 & 16.39  & 34.54 & 42.53     & 22.03      & 32.83    \\
MBR\textsubscript{BLEURT}           & 33.10 & \textbf{22.00}  & 33.01 & 42.71     & \textbf{25.31}      & 32.45    \\ 
MBR\textsubscript{COMET}           & 42.53 & 17.78  & 34.55 & 47.10     & 23.06      & 32.93    \\\hline
MBR\textsubscript{COMET}+LP         & 41.60 & 17.89  & 34.91 & 46.69     & 22.89      & 33.08    \\
MBR\textsubscript{COMET}+LP+BT      & \textbf{43.64} & 18.86  & 35.24 & \textbf{47.67}     & 23.57      & 33.17    \\
MBR\textsubscript{COMET}+LP+QE     & 42.04 & 19.96  & 35.62 & 46.89     & 23.57      & 33.76    \\
MBR\textsubscript{COMET}+LP+LM      & 41.75 & 18.40  & 35.49 & 47.56     & 23.91      & 33.85    \\ \hline
MBR\textsubscript{COMET}+LP+entropy & 42.04 & 18.34  & 35.24 & 46.24     & 22.99      & 33.16    \\
MBR\textsubscript{COMET}+LP+dropout & 41.47 & 17.90  & 34.95 & 47.43     & 22.91      & 33.10       \\ \hline
MBR\textsubscript{COMET}+LP+QE+LM & 42.24
 & 20.60 & \textbf{36.19} & 47.34     & 25.18     & \textbf{34.29}      \\
\bottomrule
\end{tabular}
}
\caption{BLEU, COMET, and BLEURT score comparison. All candidates are obtained by beam search.}
\label{table0}
\end{table*}

\section{Experiments}
\subsection{Experimental Settings}
In this section, we describe the datasets, NMT models, and metrics used in our experiments to investigate the effect of the proposed reranking methods on the $n$-best candidate list. 
\subsubsection{Datasets and Models}
To implement the NMT task, we use the German-English (De$\to$En) from IWSLT’14 task, German-English (De$\to$En), English-German (En$\to$De), and English-French (En$\to$Fr) from the WMT’14 translation task. For IWSLT’14 task, we use the data pre-processing scripts and hyperparameter settings provided by fairseq NMT repository\footnote{ \url{https://github.com/pytorch/fairseq/tree/master/examples/translation}.}. For WMT’14 task, we train a Transformer base model \cite{48} as the base NMT model and use the Newstest’14 dataset as the test set.
\subsubsection{Evaluation Metrics}
In our experiments, three widely used automatic evaluation metrics are utilized to evaluate the machine translation: BLEU, an n-gram-based precision metric which measures the lexical similarly between translation and reference; COMET \cite{27}, a multilingual and adaptable MT evaluation model, which exploits information from both source sentence and target sentence to measures the semantic similarity between translation and reference; and BLEURT \cite{28}, a learned evaluation metric based on BERT, which measures the semantic similarity between two sequences.

\subsection{Baselines}
\label{baseline}
We take the top-1 results of the beam search with beam size 5 as the baseline, which is the most widely used setting of NMT models. For all reranking methods, we follow previous work \cite{30} using beam search with beam size 30 to generate the candidates (experimental results with varying beam size and different decoding method can be found in Sec~\S\ref{sec:beamsize} and Sec~\S\ref{sec:decode}, respectively).  MBR\textsubscript{COMET} denotes use only MBR score to rank the candidate without any regularizer, where COMET is used as the utility function. Besides, we also compare MBR\textsubscript{BLEU} and MBR\textsubscript{BLEURT} which use BLEURT and BLEU as utility function, respectively. We further compare the performance of introducing different regularizer on MBR\textsubscript{COMET}, including four kinds of quality regularizer scores: log-probability (LP) score, language model (LM) score, back-translation (BT) score, quality estimation (QE) score, and two uncertainty regularizer scores: entropy score and MC-dropout score. We use GPT-2\textsubscript{base} model \cite{51} to calculate LM score. BT score and QE score is computed via backward NMT models and TransQuest \cite{50}, respectively. For the proposed method, we compute MBR score for each candidate by comparing it to partial top candidates, where the details are reported in \textbf{Appendix \ref{partial candidates}}. We also compare the method Range Voting~\cite{29} and MBR\textsubscript{BLEU}(full) \cite{20}, which using BLEU as utility function of MBR. The only difference between MBR\textsubscript{BLEU}(full) \cite{20} and our MBR\textsubscript{BLEU} is that MBR\textsubscript{BLEU}(full) uses all candidates to calculate MBR score.

%we compare LP$+$MBR\textsubscript{COMET} uses MBR score regularized by log-probability score (LP) to rank the candidates. 

\subsection{Results}
%In this section, the 1-best candidates by the proposed reranking methods are empirically analyzed. The main goals are to investigate whether the proposed reranking method is indeed capable of finding the desired output. 
We first report the results on IWSLT’14 De$\to$En and WMT’14 De$\to$En tasks. From Table \ref{table0}, we can see that MBR\textsubscript{COMET} performs better than MBR\textsubscript{BLEU}, top-1, and other baselines on all three evaluation metrics. Interestingly, we find that MBR\textsubscript{BLEURT} achieves the highest BLEURT score but low BLEU and COMET scores. To find out which utility function is the best, we further perform human evaluation (see Sec~\S\ref{sec:human}) to more quantitatively compare the reranked 1-best candidates. The human evaluation results show that MBR\textsubscript{COMET} outperforms MBR\textsubscript{BLEU} and MBR\textsubscript{BLEURT}, demonstrating that semantic-based MBR outperforms traditional lexical-based MBR. 
For the proposed regularizers, it can be found that MBR\textsubscript{COMET}$+$LP significantly improves the scores in BLEU comparing to MBR\textsubscript{COMET}. %, but not in COMET, which means that MBR\textsubscript{COMET}$+$LP ranker achieves a balance between BLEU and COMET scores by sacrificing some accuracy on COMET scores. 
Besides, MBR\textsubscript{COMET}$+$LP can be further improved in three metrics by adding other regularizers. For example, the MBR\textsubscript{COMET}$+$LP$+$QE achieves higher scores on BLEU, COMET, and BLEURT. In addition, a similar trend is observed in MBR\textsubscript{BLEURT} and MBR\textsubscript{COMET}. More results and details can be found in \textbf{Appendix \ref{utility function}}. The regularized MBR reranking has a significant improvement over the results of beam search with sizes 5 and 30, bringing 8 points and 1.5 points of improvement on COMET and BLEU metrics, respectively.

We additionally explore the performance of combining more regularizers on MBR\textsubscript{COMET}. We collectively tune the $\lambda$ value for each of the regularizers on validation sets. We observe the results of MBR\textsubscript{COMET}$ + $LP$ + $QE$ + $LM (we use RMBR\textsubscript{COMET} to denote this setting latter) that achieves the highest BLEU score among all the combinations, improving the BLEU score more than 2 points. We also find that combining quality and uncertainty regularizers with MBR\textsubscript{COMET} can not lead to further performance gains.

\begin{table}
\centering
\scalebox{0.83}{
\begin{tabular}{ccc|ccc}
\\ \toprule
& \multicolumn{2}{c}{WMT’14 En$\to$De} & \multicolumn{2}{c}{WMT’14 En$\to$Fr} \\ \hline
Methods       & COMET    & BLEU  & COMET            & BLEU        \\ \hline
Top-1 (beam=5)    & 27.24 & 27.09 & 55.11 & 38.74 \\
Top-1 (beam=30)       & 20.32 & 26.50 & 50.31 & 38.22 \\\hline
LP+QE       & 28.10 & 27.80 & 55.39 & 39.60 \\
LP+LM       & 27.92 & 28.04 & 56.10 & 39.62 \\
LP+BT       & 27.50 & 27.75 & 56.06 & 39.70 \\\hline
MBR\textsubscript{COMET}       & 34.25 & 27.37 & 59.85 & 39.18 \\
MBR\textsubscript{BLEU}        & 26.15 & 27.30 & 53.81 & 39.17 \\\hline
MBR\textsubscript{COMET}+LP    & 31.98 & 27.93 & 57.88 & 39.58 \\
MBR\textsubscript{COMET}+LP+LM & \textbf{34.97} & 28.19 & 59.80 & 39.87 \\
MBR\textsubscript{COMET}+LP+QE & 32.71 & 28.00 & 59.83 & 39.84 \\
MBR\textsubscript{COMET}+LP+BT & 32.53 & 28.01 & \textbf{60.33} & 39.83                    \\ \hline
MBR\textsubscript{COMET}+LP+QE+LM & 32.51 & \textbf{28.40} & 59.71 & \textbf{40.15}                  \\\bottomrule       
\end{tabular}}
\caption{BLEU and COMET score comparison on WMT’14 En$\to$De and WMT’14 En$\to$Fr tasks.}
\label{table2}
\end{table}

\section{Analysis}
\begin{table}
\centering
\scalebox{0.83}{
\begin{tabular}{cccc}
\toprule
Method  &  Score \\ \hline
MBR\textsubscript{COMET} & 0.281   \\ 
MBR\textsubscript{BLEURT} & 0.129   \\ 
MBR\textsubscript{BLEU}  & 0.125    \\ 
Top-1 (beam=5) & 0.120   \\ \bottomrule
\end{tabular}}
\caption{Results of the human evaluation. The score column represents the percentage of time each reranking method is judged better across its comparisons.} %Rank1 indicates the overall ranking of three automatic metrics, and Rank2 is the ranking of the human judgment scores.}
\label{table1}
\end{table}

\subsection{Human Evaluation} \label{sec:human}
From the previous results, we observe that MBR\textsubscript{COMET} outperforms MBR\textsubscript{BLEU} and MBR\textsubscript{BLEURT} in BLEU and COMET metrics, but not in BLEURT metric. This motivated us to perform human evaluation to more quantitatively compare the reranked results. For human evaluation, we randomly select a subset of 500 source sentences from the test sets of IWSLT’14 De$\to$En. Reranking is also based on the beam search results of beam size 30. We request 3 human annotators to rank the four translations from the best to the worst. Table \ref{table1} reports the ranking results according to the Expected Wins method \cite{54}. Our observation is that the 1-best candidates reranking by MBR\textsubscript{COMET} outperforms the other three methods. We provide some examples in \textbf{Appendix \ref{qualitative analysis}}. %, which is consistent with automatic evaluation (averaged ranking among the three metrics). In addition, the ranking results by human evaluation supports the ranking according to the automatic evaluation.

\subsection{Multilingual Settings}
To further verify the effectiveness of the proposed model on non-English target translation tasks, we conduct experiments on WMT’14 En$\to$Fr and En$\to$De, where we follow the same settings in Sec~\S\ref{baseline}. Since the evaluation metric BLEURT only supports evaluation the language of English, we only report BLEU and COMET scores for En$\to$Fr and En$\to$De tasks. The results are shown in Table \ref{table2}, which is consistent with the conclusion in Table \ref{table0}. 

\begin{table}[!ht]
\centering
\scalebox{0.83}{
\begin{tabular}{cccc}
\toprule[1pt]
Methods          & COMET & BLEURT & BLEU  \\ \hline
\multicolumn{4}{c}{Beam Search (beam=30)}          \\ \hline
Top-1 (beam=30) & 34.22 & 15.99  & 34.17 \\
MBR\textsubscript{COMET}           & \textbf{42.53} & 17.78  & \textbf{34.55} \\
MBR\textsubscript{BLEU}           & 34.39 & 16.39  & 34.54 \\ 
MBR\textsubscript{BLEURT}          & 33.10 & \textbf{22.00}  & 33.01 \\ \hline
\multicolumn{4}{c}{Siblings Beam Search (beam=30)} \\ \hline
Top-1 ($n=30$) & 34.11 & 15.67  & 34.09 \\
MBR\textsubscript{COMET}           & \textbf{41.44} & 17.16  & 34.39 \\
MBR\textsubscript{BLEU}           & 33.83 & 16.04  & \textbf{34.42} \\
MBR\textsubscript{BLEURT}          & 31.78 & \textbf{21.68}  & 32.95 \\ \hline
\multicolumn{4}{c}{Ancestral Sampling ($n$=30)}   \\ \hline
Top-1 ($n=30$) & 21.37 & 10.62  & 29.33 \\
MBR\textsubscript{COMET}           & \textbf{30.44} & 13.71  & 28.27 \\
MBR\textsubscript{BLEU}            & 9.67  & 8.99   & \textbf{30.62} \\
MBR\textsubscript{BLEURT}          & 9.12  & \textbf{19.74}  & 22.81  \\ \bottomrule[1pt]
\end{tabular}}
\caption{The reranking results from 30 candidates decoded by beam search, SBS, and AS on the test sets of IWSLT’14 De$\to$En.}
\label{table3}
\end{table}

\subsection{Diverse Candidate Spaces} \label{sec:decode}
From the oracle experiments (see Fig.\ref{f1}a), we observe that deterministic decoding performs better than stochastic decoding, and sibling beam search (SBS) performs as well as beam search. To further explore the effect of diverse candidate spaces, we rerank the 30 top candidates by SBS and 30 candidates sampled by AS. As shown in Table \ref{table3}, the reranking results of the candidates decoded by SB perform slightly worse than that of beam search. For AS decoding, the scores of both top-1 candidates and reranked 1-best candidates are significantly low compared to other reranking methods. %Overall, MBR with beam search decoding performs best. 

\subsection{Effect of Beam Size} \label{sec:beamsize}
To evaluate the effectiveness of larger beam size on our proposed method, we use the RMBR\textsubscript{COMET} to rank the candidates, which performs best on average of three metrics on the IWSLT’14 De$\to$En test sets. More experimental results are reported in \textbf{Appendix \ref{large beam size}}. From Fig. \ref{f2} we can see that with increased beam sizes, there is a significant improvement for COMET, BLEURT, and BLEU scores. The results suggest that our proposed reranking method can alleviate the beam search curse and generate better translations as beam size increases.

\begin{figure}
\centering
\includegraphics[width=0.4\textwidth]{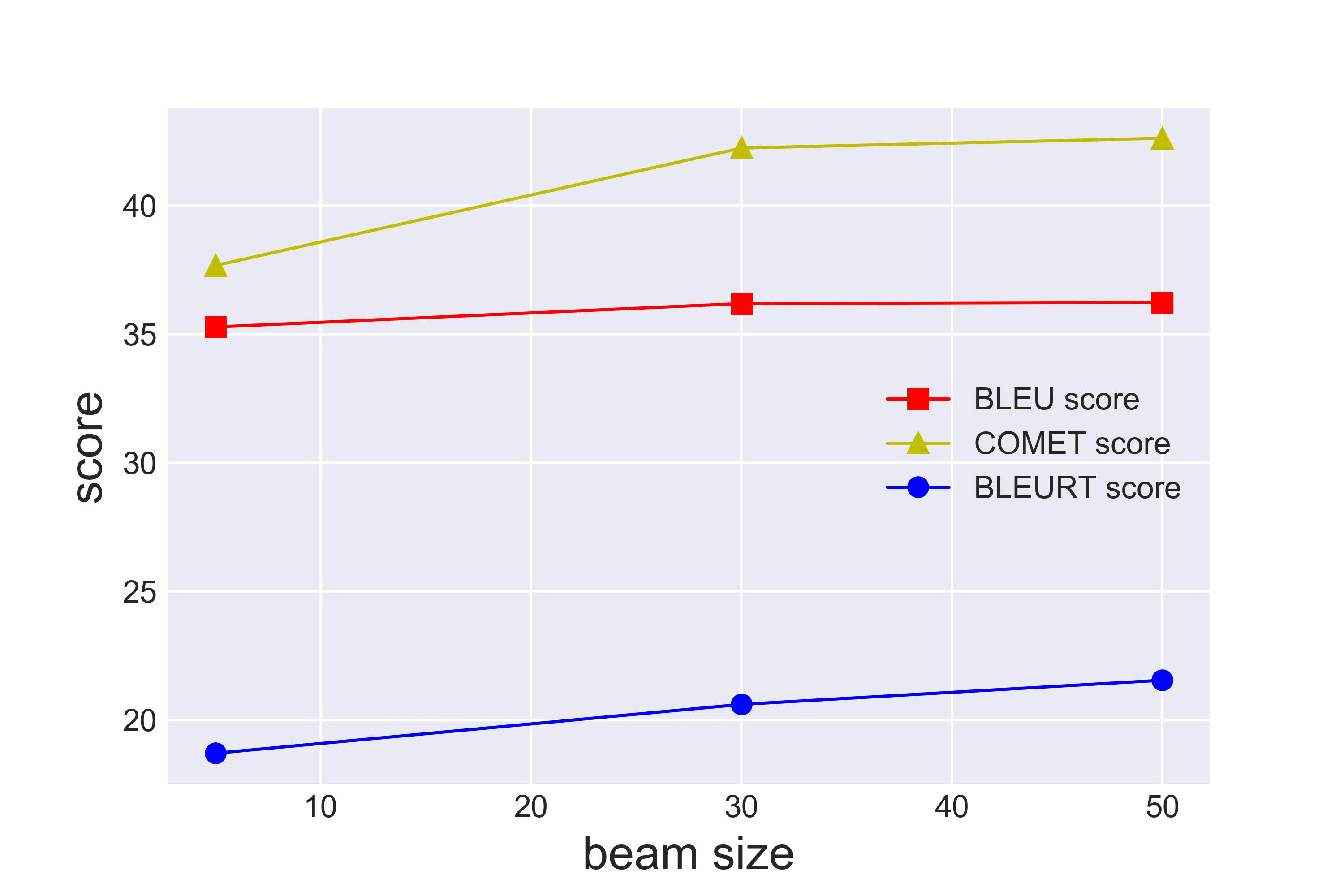}
\caption{The results of the 1-best candidates reranked by the RMBR\textsubscript{COMET} using beam of sizes 5, 30, and 50.}
\label{f2}
\end{figure}

\subsection{Inference Time}
We further compare the inference time of the proposed reranking variants and baseline. For reranking, we still use 30 candidates obtained by beam search on the IWSLT’14 De$\to$En test sets. To compare the inference time, all experiments are performs on single Tesla V100 16GB GPU. Note that, in practice we can further reduce inference time by using more GPUs to compute utility functions in parallel. The results are shown in Table \ref{table4}. $n$ represents the number of candidates used to rerank, $l$ represents the number of candidates used to compute expected utility ($n=30,l_1=21,l_2=3$). For RMBR\textsubscript{COMET}(C2F), we follow the method \cite{31} and use BLEU as the proxy utility to select 15 candidates and then use COMET as the target utility to select the 1-best candidate.
From the results we can see that RMBR\textsubscript{COMET}(n-by-l\textsubscript{1}) achieves the best performance with about 2.5 times more inference time than top-1 (beam=5). Both RMBR\textsubscript{COMET}(n-by-l\textsubscript{2}) and RMBR\textsubscript{COMET}(C2F) can further reduce inference time and outperform the baseline, which can be used as a trade-off between time cost and performance.

\begin{table}
\centering
\scalebox{0.75}{
\begin{tabular}{ccccc}
\toprule
Methods      & COMET & BLEURT & BLEU & Time \\ \hline
Top-1 (beam=5)    & 34.79 & 16.16  & 34.28  & x1     \\ 
RMBR\textsubscript{COMET}(n-by-n)       & 42.52 & 20.47 & 36.01  & x4.7     \\   
RMBR\textsubscript{COMET}(n-by-l\textsubscript{1}) & 42.24 & 20.60 & 36.19  & x3.6     \\
RMBR\textsubscript{COMET}(n-by-l\textsubscript{2}) & 40.93 & 20.26 & 35.90  & x1.4  \\ 
RMBR\textsubscript{COMET}(C2F) & 41.50 & 19.41 & 35.93  & x1.9  \\ \bottomrule
\end{tabular}}
\caption{Comparison results of inference time. Reranking uses $n=30$ candidates per sample.}
\label{table4}
\end{table}

\section{Related Work}
In NMT, reranking is a way of improving translation quality by scoring and selecting a ‘preferred’ translation from a list of candidates generated by a source-to-target model. MBR decoding is one of effective method. The goal of MBR decoding is to find a consensus translation that is closest to other candidates. Some studies rerank the $n$ candidates directly sampled from the model. \citeauthor{30}~\shortcite{30} is the first to use unbiased samples from the model by ancestral sampling, to approximate hypotheses space. Aiming at keeping computational cost of estimating expected utility tractable, a coarse-to-fine MBR procedure is proposed in \citeauthor{31}~\shortcite{31}. Other studies tend to rerank the $n$ candidates decoded by beam search. In \citeauthor{14}~\shortcite{14}, both MBR scores and log-probability scores are considered at each step of decoding. \citeauthor{20}~\shortcite{20} investigates some automatic MT evaluation metrics (BLEU, BEER, and CHRF), and observes that evaluation metric plays a major role in the $n$-best reranking approach. \citeauthor{29}~\shortcite{29} designs some similarity functions to make more informative candidates receive stronger votes, thus selecting the most representative candidate. 

These previous studies only use MBR score to rank each candidate without considering source sentence and model score. In the proposed RMBR, some regularizers are utilized to rank candidates in an overall way. Different from previous works which select candidates based on only lexical similarity, we also explore the semantic similarity between candidates. The other difference is that MBR score is computed using top-$l$ candidates of the $n$-best list to avoid candidates with poor quality in the tail list and reduce the computation cost. 

Besides MBR, there are some studies focus on MT reranking. For example, \citeauthor{37}~\shortcite{37} describes using language model to rank candidates. In \citeauthor{6}~\shortcite{6}, an energy based model is trained to rank samples drawn from NMT. %, which assigns lower energy to samples with higher BLEU score.
\citeauthor{40}~\shortcite{40} predicts the observed distribution of a desired metric, \textit{e.g.}, BLEU, over the $n$-best list by training a large transformer architecture. %\citeauthor{41}~\shortcite{41} proposes a reranking approach based on the mutual information between source and target. 
Note that these methods are orthogonal to our method, and they can be theoretically used as the quality regularizer in our framework. 

Uncertainty quantification \cite{16} have been widely used in neural networks, which is usually solved by Bayesian frameworks. Because the high training cost brought by Bayesian neural networks, various approximations, such as Monte Carlo (MC) Dropout \cite{35} and model ensembling \cite{36} have been developed. In NMT, the MC dropout is used at test time, by performing several stochastic forward passes through the model. Then, the expectation or variance of the output which reflect whether the current model is confident or hesitant on the translation, is used to evaluate machine translation quality \cite{17}. On the other hand, in the image classification task, entropy based measures are used to address uncertainty quantification \cite{19}. Our uncertainty regularizers adopt similar uncertainty quantification strategies.

\section{Conclusion}
In this paper, we introduce a RMBR to choose adequate translations from the candidates decoded by beam search. Based on MBR, we adopt semantic-based similarity and compute the expected utility for each candidate by truncating the list. The proposed quality regularizer and uncertainty regularizer are further incorporated into the framework. Extensive experimental results show that RMBR outperforms several MBR-based variants and other reranking baselines on MT tasks: +1.9 BLEU points, +7.5 COMET points, +4.4 BLEURT points over the results of beam search with sizes 5 on IWSLT’14 German$\to$English. To get a better insight into RMBR, we also conduct the in-depth ablation study and analytical experiments to show the performance improvement brought by each component of RMBR. %Through experiments, we have found that the proposed RMBR can improve upon beam search and it benefits from larger beam sizes. Besides, semantical-based MBR ranker performs better than lexic-based MBR ranker. Still, the performance gap between the best results among our proposed rankers, and oracle translations is significant. Thus, design more efficient utility functions and qualified regularizers is the target of our future work to explore better translations from the candidate space. 
%% The file named.bst is a bibliography style file for BibTeX 0.9

\bibliographystyle{named}
\bibliography{ijcai22}

\appendix

%\section{Appendix}

\section{N-by-L} 
\label{partial candidates}
\begin{figure}[!htbp]
	\centering 
    \includegraphics[width=0.4\textwidth]{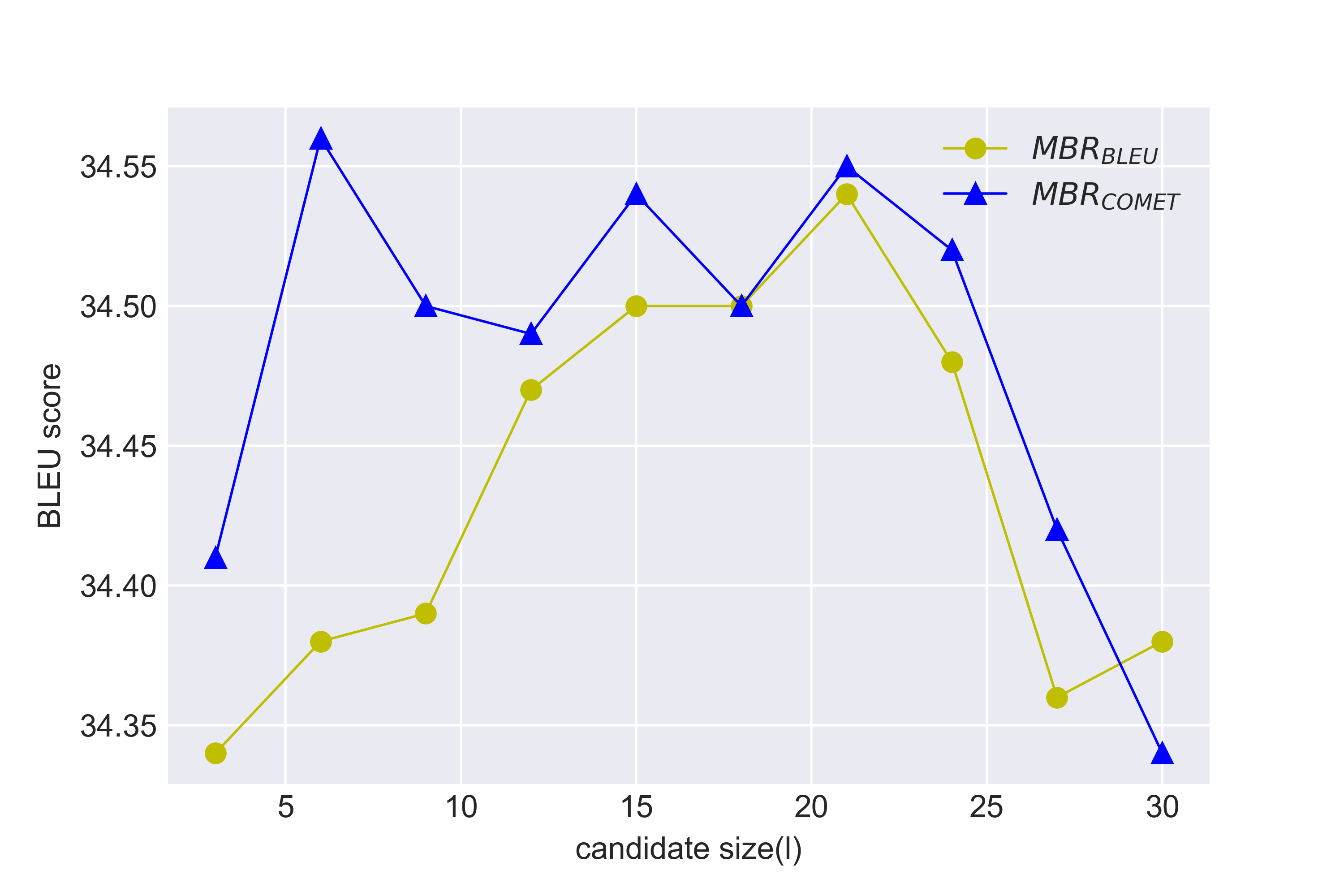}
	\caption{The reranking results using partial candidates to compute expected utility on the IWSLT’14 De$\to$En dev sets. y-axis is the BLEU score. x-axis is the number of candidates used to compute MBR scores.}
	\label{f3}
\end{figure}

The number of candidates used to compute expected utility is defined as $l$ in Sec~\S\ref{MBR Score}. To explore the effectiveness of $l$ on BLEU score of the reranked 1-best candidates, we use MBR\textsubscript{COMET} and MBR\textsubscript{BLEU} to rank the 30 candidates decoded by beam search with beam size of 30. We compute the expected utility for each candidate by comparing it to top-$l$ candidates of the 30 candidates. The results are shown in Fig. \ref{f3}. As $l$ increases, the BLEU scores of the 1-best candidates reranked by both MBR\textsubscript{COMET} and MBR\textsubscript{BLEU} go up and then down. The reason may be that partial candidates near the end of the list is extremely close to each other, but of poor quality. When $l$ increases, this part of candidates are more likely to be selected. When $l$ is around 21, BLEU scores of MBR\textsubscript{COMET} and MBR\textsubscript{BLEU} are close to the optimal. For the proposed reranking method, $l$ is tuned on the validation set and fixed for inference for all testing instances.

\begin{table}
\centering
\scalebox{0.85}{
\begin{tabular}{cccc}
\toprule
Method          & COMET & BLEURT & BLEU  \\ \hline
Top-1 (beam=5)  & 34.79 & 16.16  & 34.28 \\
Top-1 (beam=30) & 34.22 & 15.99  & 34.17 \\ \hline
MBR\textsubscript{BLEU}            & 34.39 & 16.39  & 34.54 \\
MBR\textsubscript{BLEU}+LP         & 34.75 & 16.64  & 34.56 \\
MBR\textsubscript{BLEU}+LP+BT      & 42.48 & 19.03  & 35.17 \\
MBR\textsubscript{BLEU}+LP+QE      & 38.68 & 19.75  & 35.44 \\
MBR\textsubscript{BLEU}+LP+LM      & 38.89 & 19.91  & 35.41 \\ 
MBR\textsubscript{BLEU}+LP+QE+LM      & 39.82 & 19.92  & 35.81 \\\hline
MBR\textsubscript{BLEURT}          & 33.10 & \textbf{22.00}  & 33.01 \\
MBR\textsubscript{BLEURT}+LP       & 35.83 & 19.86  & 34.55 \\
MBR\textsubscript{BLEURT}+LP+BT    & 42.46 & 19.20  & 35.18 \\
MBR\textsubscript{BLEURT}+LP+QE    & 38.91 & 20.19  & 35.42 \\
MBR\textsubscript{BLEURT}+LP+LM    & 36.79 & 18.04  & 35.25 \\
MBR\textsubscript{BLEURT}+LP+QE+LM    & 40.65 & 20.49  & 36.14\\ \hline
MBR\textsubscript{COMET}           & 42.53 & 17.78  & 34.55   \\
MBR\textsubscript{COMET}+LP         & 41.60 & 17.89  & 34.91   \\
MBR\textsubscript{COMET}+LP+BT      & \textbf{43.64} & 18.86  & 35.24    \\
MBR\textsubscript{COMET}+LP+QE     & 42.04 & 19.96  & 35.62  \\
MBR\textsubscript{COMET}+LP+LM      & 41.75 & 18.40  & 35.49    \\
MBR\textsubscript{COMET}+LP+QE+LM    & 42.24 & 20.60  & \textbf{36.19}\\\bottomrule
\end{tabular}}
\caption{Comparison results of MBR\textsubscript{BLEURT} and MBR\textsubscript{BLEU} with the proposed quality regularizers on IWSLT’14 De$\to$En.}
\label{table6}
\end{table}

\section{Utility Functions}
\label{utility function}
To further verify the effectiveness of different utility functions, we also compare the performance of introducing the quality regularizers that performs well in previous experiments on MBR\textsubscript{BLEURT} and MBR\textsubscript{BLEURT}. We follow the same settings in Sec~\S\ref{baseline}. As shown in Table \ref{table6}, similar to RMBR\textsubscript{COMET}, RMBR\textsubscript{BLEU} and RMBR\textsubscript{BLEURT} also achieve significant gains over the results of beam search with sizes 5 and 30, which is consistent with the results shown in Table \ref{table0} and Table \ref{table2}. Overall, RMBR\textsubscript{BLEURT} variants achieve better scores than RMBR\textsubscript{BLEU} variants, and RMBR\textsubscript{COMET} variants perform best. These results show that semantic-based MBR leads to better translation options.

\begin{table}
\centering
\scalebox{0.85}{
\begin{tabular}{cccc}
\toprule
Method         & COMET & BLEURT & BLEU  \\ \hline
Top-1 (beam=5) & 34.79 & 16.16  & 34.28 \\ 
Top-1 (beam=30) & 34.22 & 15.99  & 34.17 \\
Top-1 (beam=50) & 33.84 & 15.87  & 34.10 \\ \hline
\multicolumn{4}{c}{beam=50}                \\ \hline
MBR\textsubscript{COMET}          & 43.50 & 18.27  & 34.57 \\
MBR\textsubscript{COMET}+LP       & 42.35 & 18.11  & 34.94 \\
MBR\textsubscript{COMET}+LP+BT    & \textbf{44.42} & 18.97  & 35.31 \\
MBR\textsubscript{COMET}+LP+QE    & 42.74 & 20.26 & 35.62 \\
MBR\textsubscript{COMET}+LP+LM    & 42.87 & 18.96  & 35.58 \\
MBR\textsubscript{COMET}+LP+QE+LM    & 42.62 & \textbf{21.54}  & \textbf{36.24} \\ \hline
\multicolumn{4}{c}{beam=30}                 \\ \hline
MBR\textsubscript{COMET}           & 42.53 & 17.78  & 34.55  \\
MBR\textsubscript{COMET}+LP         & 41.60 & 17.89  & 34.91   \\
MBR\textsubscript{COMET}+LP+BT      & 43.64 & 18.86  & 35.24   \\
MBR\textsubscript{COMET}+LP+QE     & 42.04 & 19.96  & 35.62  \\
MBR\textsubscript{COMET}+LP+LM      & 41.75 & 18.40  & 35.49   \\ 
MBR\textsubscript{COMET}+LP+QE+LM & 42.24 & 20.60 & 36.19  \\\hline
\multicolumn{4}{c}{beam=5}                 \\ \hline
MBR\textsubscript{COMET}          & 36.65 & 16.03  & 34.19 \\
MBR\textsubscript{COMET}+LP       & 36.44 & 16.47  & 34.40 \\
MBR\textsubscript{COMET}+LP+BT    & 38.99 & 17.38  & 34.69 \\
MBR\textsubscript{COMET}+LP+QE    & 38.09 & 18.20  & 34.86 \\  
MBR\textsubscript{COMET}+LP+LM    & 36.73 & 16.81  & 34.78 \\
MBR\textsubscript{COMET}+LP+QE+LM    & 37.67 & 18.70  & 35.28 \\\bottomrule
\end{tabular}}
\caption{Comparison results of beam size 5, 30, and 50 on IWSLT’14 De$\to$En.}
\label{table7}
\end{table}

\begin{table}
\small
\centering
\begin{tabular}{l|l}
\toprule
Source    & \makecell[l]{Wir erwarten ein paar außergewöhnliche \\ Jahrzehnte.}    \\ \hline
Reference & \makecell[l]{We are living into extraordinary decades \\ ahead.} \\ \hline
Top-1 (beam=5)      & \makecell[l]{We expect some extraordinary years.}\\\hline
MBR\textsubscript{COMET} & \makecell[l]{We are \underline{looking forward to}  extraordinary \\ \underline{decades}.}    \\ \hline
MBR\textsubscript{BLEURT}     & \makecell[l]{We expect some extraordinary  \underline{decades}.}    \\  \hline
MBR\textsubscript{BLEU}      & \makecell[l]{We expect for several  \underline{remarkable}  \underline{decades}.}    \\ \bottomrule
\end{tabular}
\caption{Examples of 1-best candidates chosen by the proposed reranking methods from $n$-best list (with $n$ = 30). \underline{Underline} represents the main differences between the reference, the top-1 candidates, and the reranked 1-best candidates.}
\label{table5}
\end{table}

\section{Qualitative Analysis}
\label{qualitative analysis}
In Table \ref{table5}, we illustrate some examples from the reranking approach. Although, the word overlap between the 1-best candidates by regularized MBR ranker and the top-1 candidates is high, the proposed reranking methods produce accurate and fluent translation with asyntactic re-orderings, new words, morphological variations. 

\section{Beam Sizes}
\label{large beam size}
In this section, we explore the performance of the proposed RMBR\textsubscript{COMET} reranking in large beam sizes. As shown in Table \ref{table7}, the translation quality of beam search deceases with increased beam sizes. Notably, RMBR\textsubscript{COMET} achieves significant higher score in COMET, BLEU, and BLEURT score with larger beam size, which suggests that RMBR benefits from larger beam sizes. Moreover, the 1-best candidates of RMBR\textsubscript{COMET} far outperforms the top-1 candidates of beam search with sizes 5, 30, and 50. The results means that the proposed reranking method can improve upon beam search.

\end{document}

\section{Supplementary Material}
\subsection{Effect of Different Parts of N-best Candidates}
\begin{figure}[htbp]
	\centering 
    \includegraphics[width=0.5\textwidth]{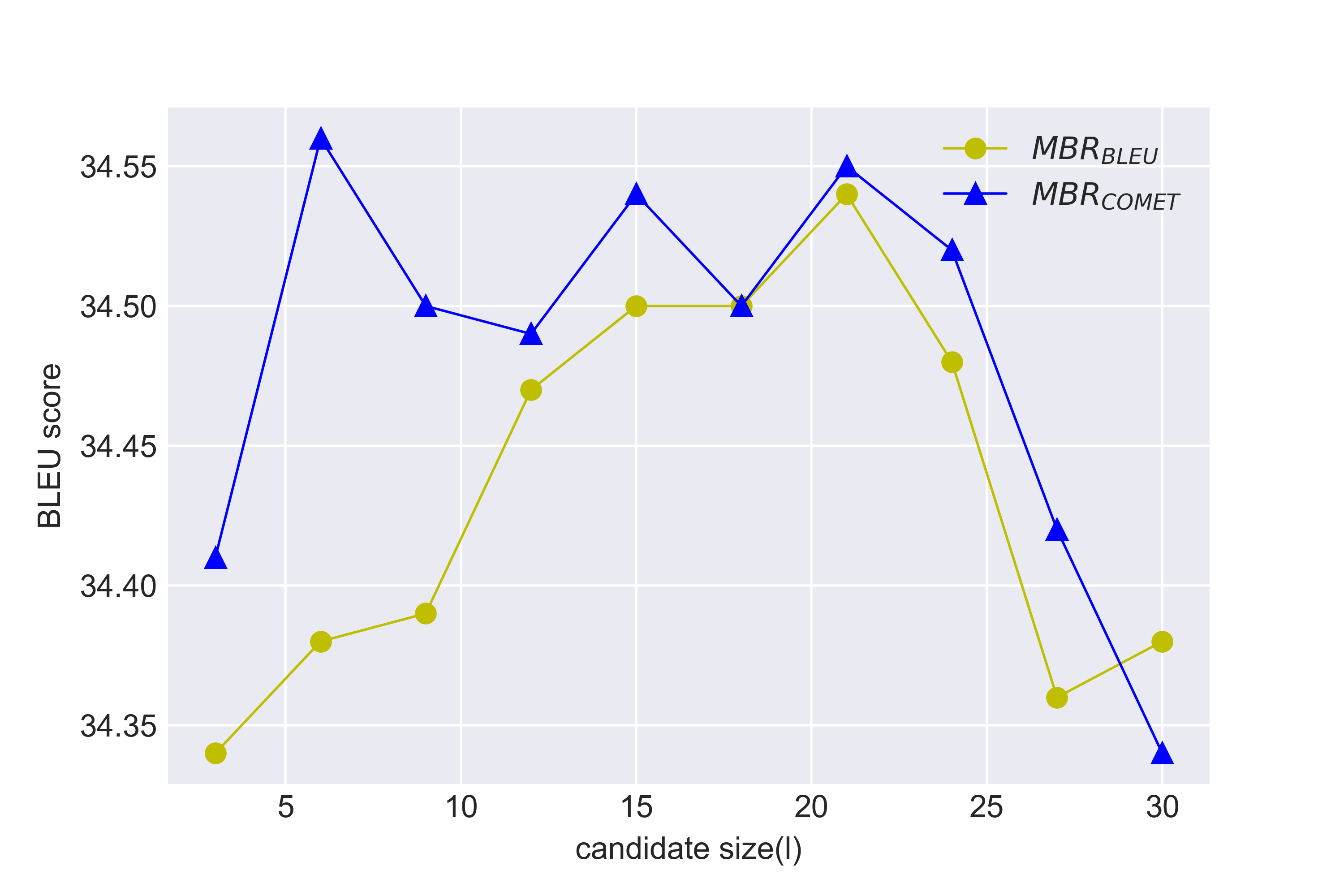}
	\caption{The quality of the 1-best candidates selected by MBR scores using different parts of candidates. y-axis is the BLEU score. x-axis is the proportion of candidates used to compute MBR scores.}
	\label{f3}
\end{figure}

We explore the effect of truncating the $n$-best list and using partial candidates to calculate MBR scores. The results are shown in Fig. \ref{f3}. Sample ratio that is defined as $r$ represents the proportion of candidates used to compute expected utility for each candidate. As $r$ increases, the BLEU scores of the 1-best candidates by both BLEU and COMET ranker go up and then down. The reason may be that partial candidates near the end of the list is extremely close to each other, but of poor quality. When $r$ increases, this part of candidates are more likely to be selected. When $r$ is around 0.7, BLEU scores of ranker BLEU and ranker COMET are close to the optimal.

\subsection{Effect of Different Utility Functions}
In Table \ref{table6}, we also report the results of using BLEU and BLEURT as the utility function to compute MBR scores. MBR\textsubscript{BLEU} ranker achieves a BLEU score of 34.54, which is still a gap from the highest BLEU score of 35.44. There is a gain in BLEU, COMET, and BLEURT scores for MBR\textsubscript{BLEU} $+$ LP or BLEURT $+$ LP ranker regularized by the quality regularizer, which is consistent with the results shown earlier.

\begin{table}[!htbp]
\centering
\begin{tabular}{cccc}
\toprule
Method          & COMET & BLEURT & BLEU  \\ \hline
Top-1 (beam=5)  & 34.22 & 15.99  & 34.17 \\
Top-1 (beam=30) & 34.79 & 16.16  & 34.28 \\
LP+BT           & \textbf{42.63} & 18.57  & 35.11 \\
LP+QE           & 38.84 & 19.53  & 35.37 \\
LP+LM           & 36.33 & 16.58  & 35.14 \\ \hline
MBR\textsubscript{BLEU}            & 34.39 & 16.39  & 34.54 \\
MBR\textsubscript{BLEU}+LP         & 34.75 & 16.64  & 34.56 \\
MBR\textsubscript{BLEU}+LP+BT      & 42.48 & 19.03  & 35.17 \\
MBR\textsubscript{BLEU}+LP+QE      & 38.68 & 19.75  & \textbf{35.44} \\
MBR\textsubscript{BLEU}+LP+LM      & 38.89 & 19.91  & 35.41 \\ \hline
MBR\textsubscript{BLEURT}          & 33.10 & \textbf{22.00}  & 33.01 \\
MBR\textsubscript{BLEURT}+LP       & 35.83 & 19.86  & 34.55 \\
MBR\textsubscript{BLEURT}+LP+BT    & 42.46 & 19.20  & 35.18 \\
MBR\textsubscript{BLEURT}+LP+QE    & 38.91 & 20.19  & 35.42 \\
MBR\textsubscript{BLEURT}+LP+LM    & 36.79 & 18.04  & 35.25 \\ \bottomrule
\end{tabular}
\caption{BLEU, COMET, and BLEURT score comparison.}
\label{table6}
\end{table}

\subsection{Results in Different Beam Sizes}
As shown in Table \ref{table7}.  There is a significant improvement for COMET scores and BLEURT scores, and a overall increasing trend for BLEU scores with increased beam sizes, which is reported in Fig. . The results suggest that our proposed re-ranking method can generate better translations as beam size increases.  There is a significant improvement for COMET scores and BLEURT scores, and a overall increasing trend for BLEU scores with increased beam sizes, which is reported in . The results suggest that our proposed re-ranking method can generate better translations as beam size increases. There is a significant improvement for COMET scores and BLEURT scores, and a overall increasing trend for BLEU scores with increased beam sizes, which is reported in Fig. . The results suggest that our proposed re-ranking method can generate better translations as beam size increases.

\begin{table}[!htbp]
\centering
\begin{tabular}{cccc}
\toprule
Method         & COMET & BLEURT & BLEU  \\ \hline
\multicolumn{4}{c}{beam=50}                \\ \hline
BT+LP          & 43.33 & 18.74  & 35.19 \\
QE+LP          & 39.12 & 19.64  & 35.35 \\
LM +LP         & 36.58 & 16.67  & 35.16 \\
MBR\textsubscript{COMET}          & 43.50 & 18.27  & 34.57 \\
MBR\textsubscript{COMET}+LP       & 42.35 & 18.11  & 34.94 \\
MBR\textsubscript{COMET}+LP+BT    & \textbf{44.42} & 18.97  & 35.31 \\
MBR\textsubscript{COMET}+LP+LM    & 42.87 & 18.96  & 35.58 \\
MBR\textsubscript{COMET}+LP+QE    & 42.74 & \textbf{20.26}  & \textbf{35.62} \\ \hline
\multicolumn{4}{c}{beam=5}                 \\ \hline
BT+LP          & 38.72 & 17.26  & 34.61 \\
QE+LP          & 36.85 & 18.26  & 34.77 \\
LM +LP         & 35.13 & 16.28  & 34.68 \\
MBR\textsubscript{COMET}          & 36.65 & 16.03  & 34.19 \\
MBR\textsubscript{COMET}+LP       & 36.44 & 16.47  & 34.40 \\
MBR\textsubscript{COMET}+LP+BT    & 38.99 & 17.38  & 34.69 \\
MBR\textsubscript{COMET}+LP+LM    & 36.73 & 16.81  & 34.78 \\
MBR\textsubscript{COMET}+LP+QE    & 38.09 & 18.20  & 34.86 \\ \hline
Top-1 (beam=5) & 34.22 & 15.99  & 34.17 \\  \bottomrule
\end{tabular}
\caption{BLEU, COMET, and BLEURT score comparison.}
\label{table7}
\end{table}
 
%% The file named.bst is a bibliography style file for BibTeX 0.9